\documentclass[letterpaper, 10 pt, conference]{ieeeconf}  % Comment this line out
\usepackage{filecontents}
\usepackage[noadjust]{cite}
\IEEEoverridecommandlockouts                              % This command is only
\usepackage{graphics} % for pdf, bitmapped graphics files
\usepackage{epsfig} % for postscript graphics files
\usepackage{mathptmx} % assumes new font selection scheme installed
\usepackage{times} % assumes new font selection scheme installed
\usepackage{tikz}

\newcommand\copyrighttext{%
  \footnotesize \textbf{Sub version. Accepted at 2018 IEEE International Conference on Intelligent Transportation Systems}\\ 
\copyright 2018 IEEE. Personal use of this material is permitted. Permission from IEEE must be obtained for all other uses, in any current or future media,
including reprinting/republishing this material for advertising or promotional purposes, creating new collective works, for resale or redistribution to servers
or lists, or reuse of any copyrighted component of this work in other works.}
\newcommand\copyrightnotice{%
\begin{tikzpicture}[remember picture,overlay]
\node[anchor=south,yshift=10pt] at (current page.south) {\fbox{\parbox{\dimexpr\textwidth-\fboxsep-\fboxrule\relax}{\copyrighttext}}};
\end{tikzpicture}%
}

\title{\LARGE {\bf{
A Look at Motion Planning for AVs at an Intersection    }} }

\author{Shravan Krishnan, Govind Aadithya R, Rahul Ramakrishnan , Vijay Arvindh and Sivanathan K % <-this % stops a space
\thanks{This work was not supported by any organization}% <-this % stops a space
\thanks{Authors are with Autonomous Systems Lab, Department of Mechatronics Engineering, SRM Institute of Science and Technology, India 
        {\tt\small shravan\_krishnan@srmuniv.edu.in}}%
}

\begin{document}

\maketitle
\thispagestyle{empty}
\pagestyle{empty}

%%%%%%%%%%%%%%%%%%%%%%%%%%%%%%%%%%%%%%%%%%%%%%%%%%%%%%%%%%%%%%%%%%%%%%%%%%%%%%%%
\begin{abstract}
Autonomous Vehicles are currently being tested in a variety of scenarios. As we move towards Autonomous Vehicles, how should intersections look? To answer that question, we break down an intersection management into the different conundrums and scenarios involved in the trajectory planning and current approaches to solve them. Then,
a brief analysis of current works in autonomous intersection is conducted. With a critical eye, we try to delve into the discrepancies of existing solutions while presenting some critical and important factors that have been addressed. Furthermore, open issues that have to be addressed are also emphasized. We also try to answer the question of how to benchmark intersection management algorithms by providing some factors that impact autonomous navigation at intersection.  
\end{abstract}

%%%%%%%%%%%%%%%%%%%%%%%%%%%%%%%%%%%%%%%%%%%%%%%%%%%%%%%%%%%%%%%%%%%%%%%%%%%%%%%%
\section{INTRODUCTION}
Autonomous driving is an emerging field with several significant developments in the recent past. Autonomous Vehicle(AV)s encounter a variety of different scenarios ranging from platooning, urban environments, intersections, highway driving, parking to name a few. Moreover, every scenario demands a different set of parameters and priorities from an autonomous perspective. Besides, the scenarios also pose different challenges in perception, decision making, tracking, path planning , motion planning and communication. Despite recent advancements in autonomous driving, the
navigation of a fleet of autonomous connected cars at an intersection is still an arduous task.

Motion planning techniques for AVs at an intersection have evolved with time and the recent past has seen many interesting approaches proposed to solve the same. This work is an attempt at presenting those proposed approaches for trajectory planning of AVs and how they have been adapted for autonomous intersections while also covering a variety of other approaches that attempt to achieve  autonomous navigation of  vehicles at an intersection. Motion planning techniques for autonomous intersection have to generate a trajectory for each vehicle that ensures the vehicle spends as little time as possible at the intersection  while accounting for obstacles(static and/or dynamic) in the environment, trajectory limits, actuator limits,  pedestrians, and interactions between vehicles. Thus, it is a challenging problem, which will have to account for a multitude of constraints and is fundamentally a multi-agent system in which the agents may have to cooperate or be non cooperative depending on their direction of travel and current states.

With the recent advancements in Vehicle to Vehicle (V2V), Vehicle to everything(V2X), Vehicle-to-Pedestrian (V2P), Vehicle-to-device(V2D) and Vehicle-to-grid (V2G)  protocols, the communication among vehicles has been utilized by intersection coordinators.  Many of the intersection algorithms implement a Vehicle to Intersection (V2I) communication for data transmission as it reduces the communication problem at an intersection. 

In this work, we attempt to analyze the current algorithms that have been presented for AV's navigation at an intersection. We look at the limitations of current works and then provide some of the open problems in the field of AVs at intersection.

The rest of the paper is organized as, in Section II we look at the problem of vehicles at an intersection and then in Section III the trajectory generation of AVs is presented. Section IV details Trajectory planning of AV in an obstacle filled environment. Section V provides a look at trajectory generation for mobile robots in dynamic environments and section VI details some methods for multi-robot trajectory generation. Section VII enumerates AVs at intersection and their approaches followed by which, section VIII  covers the drawbacks. Section IX analyses factors for comparing different intersection algorithms and details some open problems. Section X concludes the paper by answering the question of how should Intersections be adapted for AVs.

\copyrightnotice

\section{The Intersection Problem}

In this work we try to analyze the problem of \textit{\textbf{N}}  vehicles that are at an intersection attempting to traverse across it. The intersection is defined as  \textit{\textbf{M}} different roads with an arbitrary \textit{\textbf{I}} different lanes connecting across each other. We also assume that at the intersection there may or may not be any obstacles.

The aim of any vehicle is to traverse across the intersection in minimum time while also ensuring that none of the vehicles collides with other vehicles and/or obstacles, while also ensuring that the passengers' comfort in the vehicles.

%\begin{figure}
%      \centering
%      \includegraphics[scale=0.17]{Intersection1.png}
%      \caption{\bf An Autonomous Intersection consisting of four vehicles at an intersection %with the possible paths for the vehicles \cite{PODMP} }
%      \label{intersection1}
%      \end{figure}

\section{Trajectory generation for AVs}
AV can fundamentally be seen as a subset of ground based mobile robots that traverse in an ever changing and dynamic environment at higher velocities than mobile robots and are considerably larger. Trajectory generation for mobile robots is generally done by utilizing means like  RRT, Polynomials splines, A\textsuperscript{*} algorithms, Djikstra\cite{dijkstra}. In \cite{schouwenaars2002safe}, a Mixed Integer Linear Program(MILP) based method was proposed wherein the vehicle was modelled as a box which was emulated by integer constraints. They also utilized a temporal receding horizon to plan the trajectories. A polynomial based trajectory generation for mobile robots was proposed by \cite{sprunk2008planning}, who utilized quintic bezier curves to plan trajectories.

\begin{figure}
      \centering
       \includegraphics[scale=0.17]{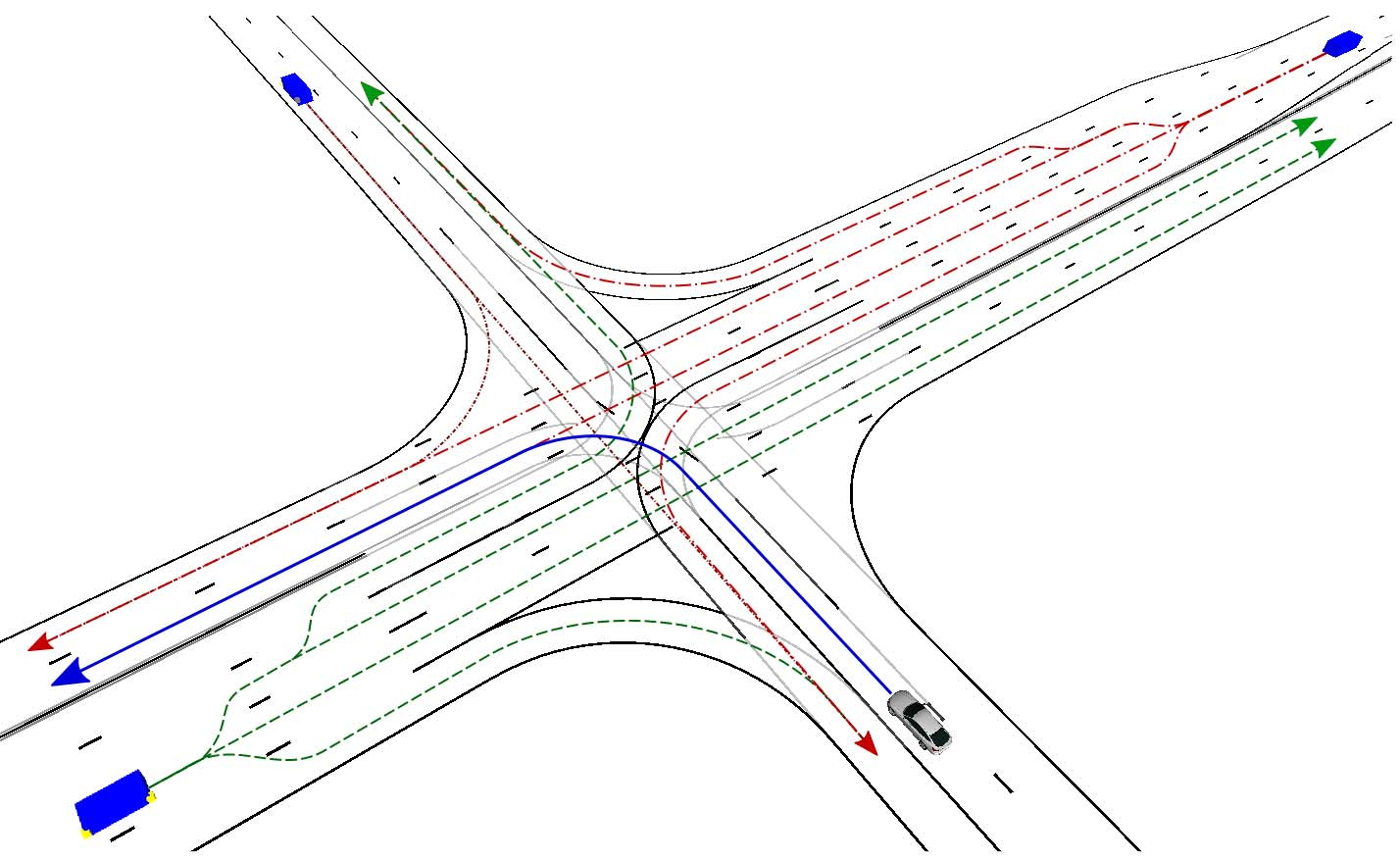}
      \caption{\bf An Autonomous Intersection consisting of four vehicles at an intersection with the possible paths for the vehicles from \cite{PODMP} }
      \label{intersection}
      \end{figure}

Trajectory planning for AVs is complicated by the bigger size of vehicles, which thereby means that dynamics is considerably more complex. One of the earliest works that attempted to achieve autonomous navigation 
 was \cite{dolgov2010path}  wherein they utilised a A\textsuperscript{*} for discrete trajectory planning, following which they utilised a Non Linear Program for smoothing the trajectory with the optimisation problem's objective consisting of distance between obstacle and vehicle, change in velocity and change in tangential angle at the vertex. This solution is locally optimal and was utilized to interpolate a cubic trajectory that is tracked. This work was one of the foremost of works that utilized optimisation techniques to generate trajectories.

High speed Trajectory planning for AVs with the dynamic constraints of the vehicle accounted was done by \cite{altche2017high} in a Model Predictive Control(MPC) formulation. This trajectory is at the near speed limits of the vehicle.

\section{Trajectory planning in Obstacle filled environment}
Trajectory planning gets even harder when it should be done for an obstacle filled environment. \cite{chomp} utilised a covariant Hamiltonian based trajectory optimisation. It was one of the first works to utilise for obstacle avoidance, an artificial potential field that consisted of the line integral parametrised arc length of the path that passed through the obstacle. This work used a Hamiltonian based monte Carlo sampling for sampling from a probability distribution. Although this work was tested for quadrupeds, it is an effective method that can be utilized for a trajectory planning for autonomous robots.

\cite{rathaionline} proposed an online trajectory re-planning and obstacle avoidance wherein they solved an 
online linear program for generating an obstacle-free lateral trajectory and then tracked the generated 
trajectory using an online linear MPC. This work while considering vehicle dynamics, assumed that longitudinal velocity of the vehicle to be constant. In \cite{plessen2017trajectory}, a linear programming based approach to plan time optimal trajectories for AVs in tube like roads was undertaken. They also incorporate the spatial dynamics of the vehicle and corridor constraints of the vehicle. This method accounts for collision avoidance constraints with the road but utilizes a kinematic based model of the vehicle.

In \cite{plessen2017}, a collision avoidance based trajectory generation for AV accounting for  geometric constraints, actuator limits and kinematic dynamics was attempted. They transform the vehicle dynamics spatially and then utilize that to generate a reference trajectory for the vehicle along the road coordinate frame and then linearize the obstacles. This optimization problem is then solved to generate a trajectory by using a Sequential Linear Program. This proposed method then provides the vehicle with linear velocity and steering commands. The system was tested in driving in a tight maneuvering scenario, thereby showing a good performance. 

In \cite{plessen2018spatial}, a spatial MPC was proposed for performing obstacle avoidance in an adaptive cruise control environment. They utilized a nonlinear bicycle model with the inputs being the throttle, braking, steering and gears. A geometric corridor is formulated by sampling at discrete points, the velocity and trajectory of the vehicle in an obstacle filled environment. Then the vehicle's trajectory is planned constraining it to be within the geometric corridor. For this work to function in a dynamic environment, continuous re-planning is required.  

Trajectory planning for AVs in presence of obstacles generally utilize a kinematic model of the vehicle to spatially transform the 
trajectory for optimizing time. The above mentioned works were all tested in simulated environments and the feasibility in a real time environment is still unknown.  

\section{Trajectory planning in Dynamic environments}
Trajectory planning in a dynamic environment is complicated by the dynamic nature of the obstacles. Thus this requires a dynamic model being utilized in these environments and the quality of the generated trajectory depends on the accuracy of the prediction model. Thus, to overcome this, a continuous re-planning of the trajectory has to be done to account for the inaccuracies in the predicted trajectory of the dynamic obstacles. The dynamic nature of the obstacles also results in possibly a non convex optimization problem.

One of the first works for trajectory planning in a dynamic environment is \cite{qu2004new} which formulated a closed form solution based on 
the kinematic model of the vehicle, dependent on the vehicle's collision probability. This solution is utilized for the formulation of 
a collision avoidance constraint based on which a trajectory is generated, differentiated twice and updated every time when the environment changes. 

In \cite{lee2017path}, a MPC based trajectory generation algorithm was presented for AVs. They utilized a trajectory planned by minimizing the distance to the desired point, jerk, velocity while also utilizing an artificial potential field for collision avoidance between the vehicles. The vehicle was considered as a box and lane constraints to the vehicle were accounted for. They tested the algorithm in simulated scenarios using other vehicles. The other vehicles' paths were simulated by a constant acceleration based model.

In \cite{gopalakrishnan2014time}, a time scaled collision cone is utilized for trajectory planning in dynamic environments. This is done by using a band of predicted trajectories and utilizing the predicted trajectories to formulate the collision avoidance. To find the nonlinear time scale, which defines the optimal trajectory of the vehicle; they utilized from the sampled solutions, the cost that is minimum of the time difference and collision.

In \cite{li2017method}, a kinematic model is utilized for optimal path planning in 2D environments. The model is transformed to a set of finite differential equations which incorporate junctions of the robots and free space. This separation and junctions are utilized to solve the problem as a stochastic differential equation based optimization. The method also solves for time and minimizes the distance traveled.

All the above mentioned works utilize a trajectory optimization based methods for generating the trajectory. In \cite{schwesinger2017motion}, sampling based approaches for motion planning are presented with two types of algorithms. The first one assumes a non cooperative motion planning between the vehicles and second a cooperative motion planning which are feasible in an environment like platooning or overtaking. The works were all tested on a real life platform. Furthermore, the thesis also experimented upon utilizing receding horizon principles for trajectory planning thereby allowing a fixed time/distance to look forward. The thesis also proposed a bounding volume based hierarchy for collision that checks for dynamic obstacle configurations if a collision looks probable and approximates continuous time trajectories from these . 

A Survey of motion planning techniques for self driving cars is presented in \cite{paden2016survey}.\cite{lefevre2014survey} puts forth a survey of the different motion prediction models. 

\section{Multi Agent Trajectory planning}
Until now we were looking at works that proceeded under the assumption of a single robot in an obstacle filled environment. An intersection is a multi-agent system with dynamic and static obstacles hence, it is pertinent to look at methods for planning trajectories for multi-agent systems.

In \cite{sutorius2017decentralized}, a decentralized planning for multi-agent systems' collaboration was proposed using polygonal based representations for non convex obstacles. They utilized a hybrid method to detect collisions and  a switching of the systems to achieve navigation.

In \cite{van2016online}, an online distributed system was presented for multiple  holonomic vehicles using Alternating Direction Method of 
Multipliers(ADMM). The Vehicle was formulated by a kinematic model and the collision avoidance constraints were enforced using separating 
hyperplanes. Then a slack of the trajectory of other vehicles is available for the ego vehicle to generate feasible collision-free trajectories. 

In \cite{plessen2016multi}, proposed a bi-directional vehicle coordination scheme 
utilizing prioritized decoupling of path planning with an aim to reach platooning of 
vehicles quickly. The path planning is done by a centralized system and the 
reference trajectories are published back to the respective vehicles.

In \cite{tang2018hold}, a centralized multi robot trajectory planner for 2D obstacle-free environments was proposed utilizing tools from nonlinear optimization and calculus of variation. Furthermore, the approach utilized a two step process for planning trajectories wherein a piecewise linear trajectory is generated based upon geometric constraints in the first step and in the subsequent step a higher order polynomial parameterized trajectory is generated. They tested the method using quadrotors in an obstacle-free environment, but it is tractable for mobile robots also.

\section{Intersection management}
While the above works presented approaches to trajectory planning for AVs in different scenarios, trajectory generation for intersection is dynamically evolving in terms of vehicle quantity, positions, and obstacles. This section attempts to review some of the works with respect to autonomous intersection management systems. 

From a high level perspective, autonomous intersection management systems can be classified based on the methods utilized as:
\begin{itemize}
\item \textbf{Multi-Agent Systems:} These management systems consider autonomous intersections as a series of multi agent navigation problems with each agent having a set goal to reach with a trajectory to be generated either as a set of time stamped poses and/or control actions. \cite{onieva2015multi}, \cite{murgovski2015convex}, \cite{gregoire2013optimal} 
\item \textbf{Slot Based Systems:} The following works consider autonomous intersections as slots provided to vehicles from different sides similar to air traffic management systems with the vehicles requesting for reservation to utilize the intersection area \cite{Light_Traffic} \cite{AIM}, \cite{qian2017autonomous}, \cite{he2018erasing}
\item \textbf{Detection Based:} These algorithms are not designed for autonomous intersection management systems but rather a particular set of scenarios like a blind intersection for a pedestrian/cyclist as presented in \cite{yoshihara2017autonomous}
\end{itemize}

\subsection{Slot Based systems}

Slot based systems are currently popular as they allow vehicles' time-slots to traverse across intersection.
\cite{AIM} proposed a multi agent approach to AVs that utilized a scheduling based system. A vehicle, upon arrival at the intersection, sends a request to the intersection manager that provides a response back detailing the velocity, trajectory and outbound lane for the vehicle after analyzing the intersection. This method utilizes a multi-agent system with the intersection split into grids and collision of vehicles checked along the grid points, resulting in a discrete time approach. An improvement over this was proposed in \cite{levin2017conflict}, where a reservation based approach for reserving space-time for a vehicle.  The algorithm bypasses the requirements to communicate and request for space. They introduced a mixed integer linear program to optimize for reservation and then using a temporal rolling horizon, they move the horizon forward at each instance. They further improved upon the discrete time based approach proposed in \cite{AIM} by utilizing continuous- time collision avoidance verification by using conflict points and checking for collision only at the conflict points.

In \cite{lu2014rule}, a set of rules for coordination at an intersection based on Chinese traffic rules(Right hand driving) was proposed that stated whether another vehicle has to yield or move in the right of way for the system. In \cite{savic2017distributed}, a distributed algorithm for intersection crossing that implemented  transmission of estimated state, uncertainty and desired lanes based on which the higher priority vehicle traversed. The messages passed are acceleration, velocity, longitudinal and lateral position. In \cite{he2018erasing}, an all direction turn lanes was proposed which employs the fact that AVs need not utilize directional turning lanes and can function optimally in scenarios that require the vehicles to turn out in different directions from different lanes due to their high rate of control. Based upon this paradigm, collision-free regions are provided for the vehicles depending on its position and other available vehicles.

\cite{levin2017optimizing} proposed a method to optimize reservation algorithms that use a tile/grid based trajectory discretization. They addressed the problem of resolving non-cooperating(conflicting) requests that have to be optimized, an integer program was put to use to solve for optimal reservation strategies.

\cite{fayazi2017} utilized a mixed integer linear program to solve for optimal scheduling of the vehicles using  bi-directional communication between the vehicles by trying to minimize the time of access for vehicles at the intersection and constraining the maximum access times for vehicles.    

In \cite{Light_Traffic},two intersection algorithms were proposed using queuing theory and a slot based system. The proposed two methods are based upon queuing theory and were formulated to balance the paradigms of vehicle flows or capacity, with each one emphasizing on either one of the paradigms. Based upon these paradigms and flows of the vehicles, time scale of intersection access is defined.

\subsection{Multi Agent approaches}

The above discussed methods utilized communication based strategies for collision avoidance which do not account for how the agents' motion changes accurately or provide an optimal path to vehicles. To overcome these drawbacks, a decentralized or a mixed system will provide better efficiency for optimal traversal of vehicles. 

In \cite{onieva2015multi}, a multi-objective evolutionary algorithm was developed to 
develop a fuzzy rule base for controlling the speed of the vehicles entering the intersection. The objective for optimization consists of the number of vehicles, the vehicles' position and velocity. They assume that the vehicles do not collaborate with each 
other. The generated trajectory consists of vehicles' speeds to traverse across the intersection. In \cite{bui2017cooperative}, a game theoretic approach to optimize traffic flow across multiple intersections was proposed.  They formulate the traffic intersection as a non-cooperative game among different players and then based upon requests from different players (vehicles), the signal lights are changed. While the above proposed method function is intended only for traffic light control, the approach however can be utilized by both AV and non AVs.

In \cite{murgovski2015convex}, the AVs' space-time trajectories are transformed to a spatial dynamic model thereby reducing the problem's dimension. Based upon this formulation, a centralized optimization problem for all vehicles available within the intersection's control radius is solved. This is easily solved as it is a convex region for collision. The objective is also reformulated spatially and penalized as a quadratic function, thereby solving a Quadratic Program. Based upon this, the authors of\cite{riegger2016centralized}
proposed a centralized MPC for autonomous intersections that utilized the spatial reformulation of dynamics and solved a Quadratic Program . They added slack variables to compensate for spatial discretization of dynamics and a collision avoidance constraint that allows only one vehicle at a temporal point in the critical region of intersection. This results in an easy collision avoidance and is not scalable for heavy volumes of traffic flow and increases the time spent at an intersection. The mentioned works also require that the vehicles have a reference trajectory throughout the intersection which may not be practical. 

Furthermore, \cite{belkhouche2017control} proposed an online optimization algorithm for controlling the vehicle speeds utilizing Minkowski addition and formulated unsignalised intersection problem with speed ratios for collision avoidance. They also used Minkowski addition to transform vehicles to points and solved a Quadratic Program for control actions. The above work was simulated for scalability but assumes control of the vehicle. Hence, it is viewed as a centralized system thus requires a huge computation as the vehicle number scales.

To overcome the problems with centralized approaches to autonomous intersection, \cite{katriniok2017distributed} proposed a distributed MPC for autonomous intersections utilizing parallel optimization. The distribution is done by giving every individual vehicle a local objective and overall constraints thereby resolving the centralized optimization problem into a quadratically constrained quadratic program. Non convex collision constraints are prioritized and converted to a semi definite programming that is solved online. \cite{hult2015approximate} proposed a combinatorial optimization problem that is decomposed into two problems, a first problem of finding the optimal coordination of the vehicles centrally and then subsequently in each vehicle a local optimal control problem is solved depending on the systems constraints and timing slots specified to each vehicle.

\cite{gregoire2013optimal} Proposed a reservation based optimization wherein the problem is broken down into two parts. The first part is used for scheduling vehicles' traversal times and then in the second step a continuous time trajectory is generated for the vehicles. They propose a heuristic to find locally optimal solutions for the second step. 

The above mentioned works required the vehicles transmit large amount of data consisting of the vehicle's current states and desired states. This raises privacy constraints and can result in failure if there is data loss or transmission failures. In the recent past, some works have been attempted to overcome that problem.

\subsection{Under data unavailability/ uncertainty}

A Mixed Observability Markov Decision Process was utilized in \cite{MODMP} as a method of intersection merging, wherein a probabilistic model, consisting of two states( move and stop) and the velocity profile of the other vehicles that are present in the environment, is utilized for the control actions. The method was tested in the presence of human drivers and their intents. Then in \cite{PODMP}, a partially observable Markov decision process was utilized with other vehicles' intentions as hidden variables.  They formulated an optimization problem for finding the optimal acceleration of vehicle along pre-planned paths.  They recast the problem to a lower dimension and solved a continuous time path planning for vehicles considering future layout and future uncertainties. 

Due to an abstraction of desired goals and current states, the above works perform utmost worse than the methods that function without considering data abstractions. 

\section{Drawbacks of the current state of the art algorithms}
Reservation based intersection management systems provide vehicles with a time slot  and a speed at which they have to travel across an intersection. This, while reducing the burden on the system, often results in a vehicle having to slow down at the intersection. Moreover, these algorithms  sample trajectories at grids or tiles, the collision avoidance's accuracy is dependent on the discretization. Besides, the first come first serve paradigm of these algorithms result in sub optimal trajectories. These algorithms albeit are much better for heterogeneous vehicles.

On the other hand, optimal control/MPC based approaches couple the problems of trajectory optimization and tracking while spatially reforming the trajectories, a  reformulation that results in time becoming a function of position and it's derivatives. These approaches are primarily efficient in a centralized system and the resultant problem's complexity scales with number of vehicles and/or vehicles with different directions of flow. These complexities result from a higher dimension space and higher interactions to be accounted for. Furthermore,  most of these algorithms were tested in a specific scenario and/or with limited number of vehicles.

Therefore, on the whole, the drawbacks of current intersection management systems are:

\begin{enumerate}
\item \textit{Re-initialization of algorithms are difficult}: This is an important problem in the  optimal control based approaches, re-initializing the solver in the next iteration appropriately with respect to either the trajectory of the previously available vehicles or utilizing previously generated trajectories for the vehicles and employ them as a starting point for next iteration when new vehicles might be added and/or some vehicles might exit the intersection. Hence, it is important to look at methods to reinitialize the numerical optimal control problem for subsequent iterations. 
\item \textit{Scalability of the algorithms}: Majority of the optimal control based formulations that use MPC based approaches have been tested with a meager amount of vehicles. This performance may not be tractable when the number of vehicles is in tens and hundreds as this adds huge computational burden.
\item \textit{Conservative/non-realistic collision approximations}:  The vehicle size is either modeled as a circle or polygon. Formulation of vehicles as polygons is much more appropriate but many prevalent algorithms model vehicles as circular, which is a conservative approximation of the vehicles' region of interest. On the other end of the spectrum, some approaches  formulate the solution allowing only a limited number of vehicles onto the intersection region. This conservatively uses a huge region and thereby misappropriates available space.  A step towards eliminating the latter has been attempted by \cite{he2018erasing}.
\item \textit{Discrete intersection regions}: The reservation and slot based approaches simulate discrete space grids and then check for vehicles' collisions at the grids. Despite recent advancements that evaluate collisions only at important points, the usage of discrete approaches always have a probability of excluding critical points of collisions. 
\item \textit{Systems developed for specific types of intersections}: The developed algorithms are implemented and tested for specific type of intersections like  four way, 'T' intersection to name a few. Moreover, some of the developed intersection managers also require previously planned collision-free trajectories. It is important that the developed algorithms perform for a wide variety of intersection configurations for an efficient performance.
\item \textit{The optimization of all control actions centrally}: The intersection managers, generally plan the overall trajectories of the vehicles in a centralized manner. These methods take up a lot of computational burden as the number of vehicles scale up. Despite recent advances in intersection managers that utilize distributed systems for optimization, they still require a central system for some tasks.
\item \textit{Robustness}: Any sensory data has some uncertainty associated with  it. The uncertainty in the data is not accounted in current algorithms. These uncertainties might result in drastic changes in the feasible region as the vehicles' uncertainty grows as they move forward.
\item \textit{Re-planning of trajectories}: Once a vehicle's trajectory is planned, majority of intersection managers do not consider the online re-planning of the trajectories, this results in wastage of space at the intersection. Re-planning of trajectory will result in better travel as an intersection is a dynamic environment and re-planning will allow better use of available space.
\end{enumerate}

\begin{table*}
\caption{Comparison of six algorithms according to the proposed factors}
\label{table_1}
\begin{center}
\begin{tabular}{|c|c|c|c|c|c|c|c|}
\hline
\textbf{Algorithm} & \textbf{Scalability} & \textbf{Time Spent} & \textbf{Smoothness} & \textbf{Usage of Space} & \textbf{Heterogeneity} & \textbf{Robustness} & \textbf{Versatility}   \\
\hline
\cite{AIM} & Excellent & Medium & Acceleration & Medium & High & Message Drop & Modifiable  \\
\hline
\cite{levin2017conflict} & Excellent & Medium to Low & Acceleration & Medium & High & Message Drop & Medium  \\
\hline
\cite{Light_Traffic} & Excellent & Low & Acceleration & Medium & High & Low & Medium  \\
\hline
\cite{riegger2016centralized} & Low & High & Jerk & Conservative & Low to Medium & No & Low   \\
\hline
\cite{katriniok2017distributed} & Medium & Low to medium & Jerk & Medium & Low to Medium & No & Medium \\
\hline
\cite{PODMP} & Low & Medium & Acceleration & Low & High & High & Medium \\
\hline
\end{tabular}

\vspace{2mm}

\begin{tabular}{|c|c|c|}
\hline
\textbf{Algorithm}   & \textbf{Data Requirement} & \textbf{Data Privacy}  \\
\hline
\cite{AIM}  & Medium & Low \\
\hline
\cite{levin2017conflict} & Low  & Limited \\
\hline
\cite{Light_Traffic} & Low  & Limited \\
\hline
\cite{riegger2016centralized} & High & Limited \\
\hline
\cite{katriniok2017distributed} & Low & Limited \\
\hline
\cite{PODMP} & Low & High \\
\hline
\end{tabular}
\end{center}
\end{table*}

Furthermore, the proposed algorithms do not take into account lane marking constraints,  obstacles, pedestrians, cyclists and other obstacles that might occur at an intersection. This means that for utilizing the actions from the developed intersection managers, a local trajectory planning/re-planning has to be done by the vehicle.

\section{Analysis and Open Problems}
\subsection{Factors to consider for Intersection}
With the recent developments in autonomous intersections, a method to compare and contrast intersection managers is important. Some factors which we think are important to analyze intersection management systems are:

\begin{enumerate}
\item \textit{Scalability}: An intersection manager should be able to handle a large number of vehicles efficiently while also showcasing a similar performance when the number of vehicles at intersection are limited.
\item \textit{Time Spent at intersection}: It is important that all vehicles spend as limited time as possible at intersections. An analysis of the time spent at intersection will provide an understanding of the performance of the intersections as it reduces overall commute times and reduces emissions and energy expenditures. 
\item \textit{Smoothness of trajectories}: The generated trajectories that the vehicles follow need to be continuous at least until the jerk of the vehicles as AVs have third order dynamics. Furthermore, a higher order of smoothness offers a higher comfort for passengers and increases lifetime of actuators.
\item \textit{Usage of intersection space}: The intersection space if necessary can be completely filled and an analysis of how the intersection space is utilized will allow intersection managers to be compared much better. Furthermore, the intersection spaces are generally a non convex space but it will still be useful to analyze algorithms by the amount of empty space left during peak handling capacities and/or the restrictions they apply for occupancy.  
\item \textit{Heterogeneity}: The ability of an intersection to handle vehicles of different sizes, dynamics, limits and constraints is an important point to analyze and compare intersection management systems. A greater ability to handle heterogeneity is important for intersection managers.  
\item \textit{Robustness}: The usage of data uncertainty by intersection managers is an important consideration for analyzing intersection algorithms. Utilizing the uncertainties will increase the possibility of collision-free trajectories for AVs during higher volumes of vehicles at the intersection. Moreover, due to forward simulation nature of autonomous intersection, the uncertainty is bound to increase and accounting for that might result in conservativeness with respect to time spent at intersections. Hence, an appropriate trade-off between the two will have to be considered and  is another method for analyzing different algorithms. 
\item \textit{Versatility}: With many of the current algorithms being developed and/or tested keeping in mind certain configurations, it is important to analyze the performance of intersection algorithms when layouts of the intersection change. 
\end{enumerate}

While the previous factors looked at analyzing algorithms from the perspective of generated trajectories and control actions, the intersection management algorithms generally require data sharing. Thus,  it is equally important to consider the working of the system from a data sharing and requirements perspective.

\begin{enumerate}
\item \textit{Data required to manage trajectories}: Some intersection managers require the vehicles to share their overall control and state trajectories to the intersection managers and this data sharing results in handling large amount of data . The handling of such large data increases the probability of data losses and the latency of communicated data. The amount of data required by an intersection manager per vehicle is also an important consideration for bench-marking intersection management algorithms. 
\item \textit{Vehicle Privacy}: With the recent concerns with respect to privacy and data theft, it is pertinent to analyze whether the intersection manager ensures privacy with respect to storage, utilization of data and the data that vehicles transmit among themselves. The reduction in the amount of data shared increases the probability of predicted trajectories deviating heavily from the actual trajectories. This makes data privacy as another essential aspect to consider for intersection management algorithms.
\end{enumerate}

A comparison of some algorithms based on the presented factors is given in Table \ref{table_1}

\subsection{Open Problems}
Autonomous Intersection is still a developing field with a multitude of algorithms developed for navigating through an intersection, some problems that still remain are:

\begin{enumerate}
\item \textit{Decentralized Algorithms for navigation}: While current algorithms for intersection managements are progressing towards distributed optimization, a completely decentralized approach will require decentralized communication among vehicles. This results in trajectory planning taking into consideration only vehicles within the vicinity. These methods will share the computational burden among vehicles and allow for individual decisions among the vehicles in manner that not only prevents collision but also prioritizes its requirements.
\item \textit{Privacy constraints not affecting solution}: Recently, intersection management algorithms have been proposed that perform with abstraction in the intentions of other vehicles in the environment. Such systems are still in their nascent stages and perform less favorably than systems that know drivers' intentions. Employing techniques for vehicle prediction and intention is an avenue for further research. 
\item \textit{Learning for different subsystems}: Learning based algorithms will provide a method to utilize experience from previous observations to aid in current maneuvers. Learning can be added in prediction of trajectories, optimization of trajectories or in assigning priorities to vehicles and their movements. This will allow a system to gain knowledge from how other vehicles react and thereby provide an optimum level to compensate for the uncertainties associated with the systems and provide safety to the systems. 
\item \textit{Obstacles in the environment}: Despite the dynamic nature of intersection and the presence of multiple different obstacles ranging from pedestrians, shops , cyclists ; Intersection algorithms tend to ignore them or compensate for them by requiring the vehicles provide a collision free reference. The consideration of such obstacles will also account for the most important aspect of a transportation management system, the safety of passengers and humans that utilize the roads for different purposes.   
\item \textit{Hybrid models for prediction}:Intersections require a method to predict the trajectories of the other vehicles and/or obstacles that are contained in the environment. These predictions account for either the dynamic nature, interaction between different vehicles and/or maneuver based predictions. Maneuver based predictions offer a good understanding of the vehicle's direction but doesn't account for the interactions possible, which can be compensated for by utilizing interactions. A combination of these different methods for prediction will allow for a more versatile and accurate prediction. This will also allow smoother evasion and/or better collision avoidance by AVs. 
\item \textit{Data uncertainty}: Data uncertainty is another important field for autonomous intersections and will increase robustness and incorporate a versatile collision avoidance among the vehicles. Adding data uncertainty into the system will also allow for intersection algorithms to incorporate real life uncertainties that are going to exist due to inaccuracies and model mismatch. Moreover, data uncertainty will allow accounting for the stochasticity with the vehicle maneuvers that are difficult to predict.
\end{enumerate}

Majority of the above proposed problems tend to overlap with solving another problem. All of these problems are important considerations for achieving zero collisions and minimum time as the vehicle traverses the intersection. An optimal trade-off among many of the above factors is required.   

\section{Conclusion}
The current state of the art intersection algorithms attempt to solve the problem 
as a reservation for the vehicles. These reservation based algorithms discretize the trajectory into tiles, and 
simulate for collisions at the tiles. These methods are not feasible in real life scenarios and may not be scalable.
A second set of algorithms try to formulate a centralized problem that aims at generating optimal trajectories for 
vehicles at an intersection utilizing techniques from optimal control, convex optimization and fuzzy systems. These 
systems have to solve an online optimization problem every time a new set of data is received by the intersection 
manager. Solving the problem as its dimension increases (number of vehicles) is increasingly 
difficult. Hence, approximation of the solution is achieved by different methods.

Revisiting the question of, how should autonomous intersections look, we believe that intersections being a dynamic environment should have continuous trajectory re-planning and utilize the full space of the intersections. The utilization of full space of the intersection necessitates the removal of lanes. Furthermore, safety of humans is of utmost importance at intersections and hence the consideration of pedestrians, cyclists and other human related obstacles are of prime importance for intersection algorithms. The recent data issues have also highlighted that privacy is of utmost concern. Thus, moving forward utilizing decentralized algorithms for trajectory re-planning and communication is of higher priority as it reduces computational burden but also ensures adherence to a degree of privacy, prioritizing and cooperative driving among the vehicles.

\addtolength{\textheight}{-0.4cm}   % This command serves to balance the column lengths
                                  % on the last page of the document manually. It shortens
                                  % the textheight of the last page by a suitable amount.
                                  % This command does not take effect until the next page
                                  % so it should come on the page before the last. Make
                                  % sure that you do not shorten the textheight too much.

\end{document}